\pdfoutput=1

\documentclass[11pt]{article}

\usepackage{EACL2023}

\usepackage{times}
\usepackage{latexsym}
\usepackage{graphicx}
\usepackage{amsmath}
\usepackage{multirow}
\usepackage{stfloats}
\usepackage{algorithmic}
\usepackage{algorithm}
\usepackage{booktabs} 
\usepackage[T1]{fontenc}

\usepackage[utf8]{inputenc}

\usepackage{microtype}

\usepackage{inconsolata}

\title{SPUQ: Perturbation-Based Uncertainty Quantification\\for Large Language Models}

\author{Xiang Gao, Jiaxin Zhang, Lalla Mouatadid, Kamalika Das \\
  Intuit AI Research \\
  2700 Coast Avenue, Mountain View, CA 94043 \\
  \texttt\{xiang\_gao, jiaxin\_zhang, lalla\_mouatadid, kamalika\_das\}@intuit.com}

\begin{document}
\maketitle
\begin{abstract}
In recent years, large language models (LLMs) have become increasingly prevalent, offering remarkable text generation capabilities. However, a pressing challenge is their tendency to make confidently wrong predictions, highlighting the critical need for uncertainty quantification (UQ) in LLMs. While previous works have mainly focused on addressing aleatoric uncertainty, the full spectrum of uncertainties, including epistemic, remains inadequately explored. Motivated by this gap, we introduce a novel UQ method, sampling with perturbation for UQ (SPUQ), designed to tackle both aleatoric and epistemic uncertainties. The method entails generating a set of perturbations for LLM inputs, sampling outputs for each perturbation, and incorporating an aggregation module that generalizes the sampling uncertainty approach for text generation tasks. Through extensive experiments on various datasets, we investigated different perturbation and aggregation techniques. Our findings show a substantial improvement in model uncertainty calibration, with a reduction in Expected Calibration Error (ECE) by 50\% on average. Our findings suggest that our proposed UQ method offers promising steps toward enhancing the reliability and trustworthiness of LLMs.
\end{abstract}

\section{Introduction}

\begin{figure}[h]
\centering
\includegraphics[width=7cm]{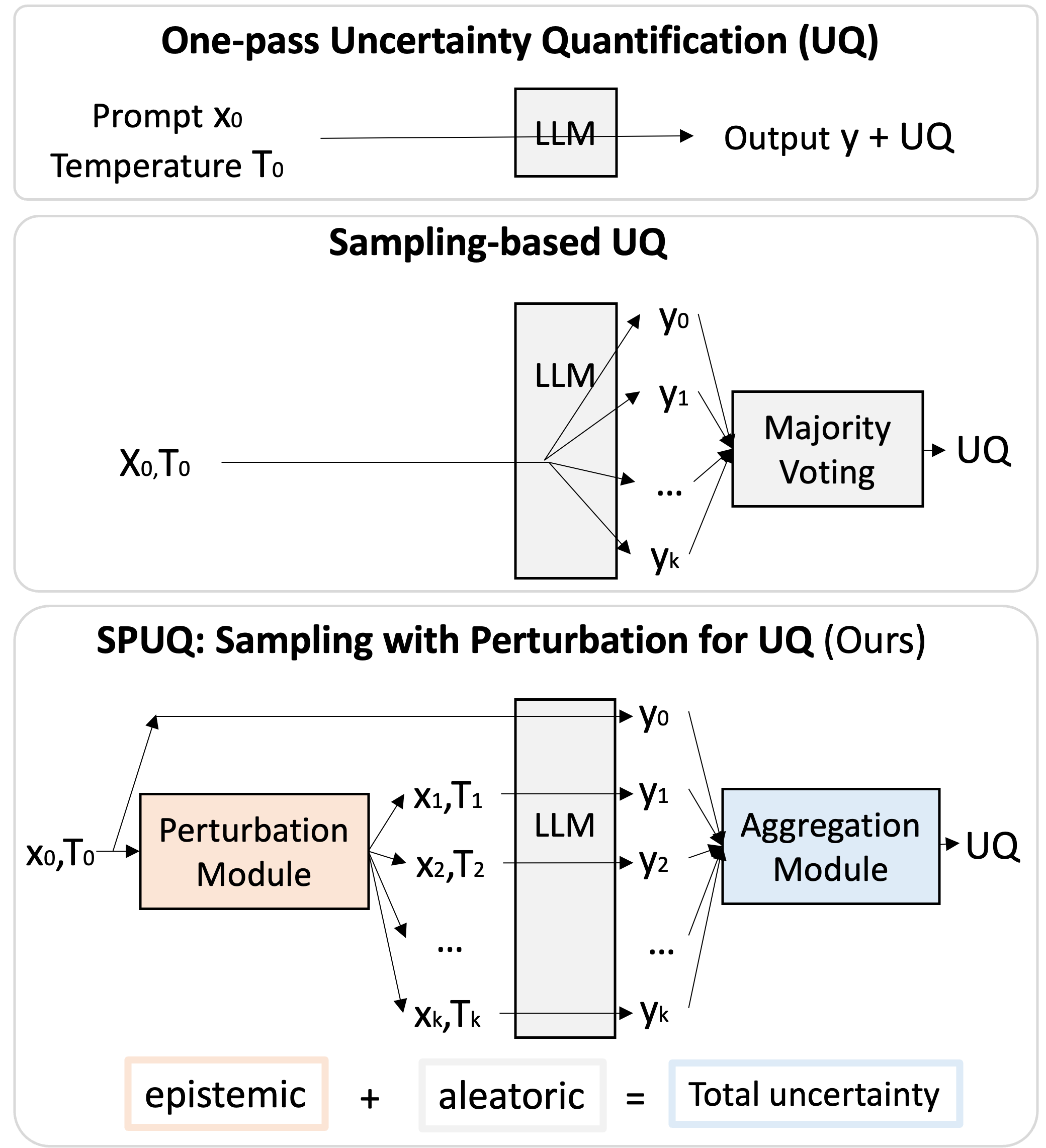}
\caption{\label{fig-flowchart}
Overview of uncertainty quantification techniques: one-pass \cite{lin2022teaching, kadavath2022language, chen1998evaluation}, sampling-based \cite{si2022prompting, wang2022self}, and our SPUQ method. SPUQ addresses both epistemic (via perturbation) and aleatoric (via sampling) uncertainties. Aggregation yields the total uncertainty, distinguishing SPUQ from traditional methods focused mainly on aleatoric uncertainty.
}
\end{figure}

Large language models (LLMs) \cite{brown2020gpt3, chowdhery2022palm, touvron2023llama, chatgpt, chung2022flan, openai2023gpt4} have demonstrated remarkable success in various natural language processing tasks, such as text generation and question answering. However, a significant concern with LLMs is their proclivity to hallucinate, or make confidently wrong predictions \cite{maynez2020faithfulness, zhang2023snowball, ji2023hallucination, chen2023robust}, even for advanced models like GPT-4 \cite{openai2023gpt4}. To address this issue, a robust approach to quantify uncertainty is necessary, enhancing the reliability and trustworthiness of these models \cite{si2022prompting, lin2022teaching, zhou2023navigating, kuhn2023semantic}. This is particularly important in online scenarios where the reference answers are not available, so that a response with a low confidence score can be treated with appropriate skepticism.

Uncertainty in machine learning models can be categorized into \emph{aleatoric} (data-wise) and \emph{epistemic} (model-wise) uncertainty \cite{hora1996aleatory, hullermeier2021aleatoric}. In the case of LLMs, \emph{aleatoric} uncertainty arises from the inherent variability of natural language, where multiple or even an infinite number of valid outputs can exist for the same input. 
Existing work has proposed methods relevant to aleatoric uncertainty. A common approach is to use the (normalized) generation likelihood (or the reciprocal of perplexity \cite{chen1998evaluation}) as an uncertainty measure. However, many popular LLMs, such as ChatGPT and GPT-4, do not provide access to token probabilities in their APIs, rendering this approach inapplicable. Additionally, sampling-based methods \cite{wang2022self, si2022prompting, kuhn2023semantic} examine the deviation among multiple outputs generated from the original input. However, when an LLM makes a confidently incorrect prediction, resampling tends to yield similar results, leading to an overconfident and poorly calibrated score.

\emph{Epistemic} uncertainty, on the other hand, is due to the lack of knowledge or suboptimal modeling. 
It captures the uncertainty that arises from limitations in the model's ability to learn from the data and generalize to new situations. Existing literature has linked epistemic uncertainty with adversarial examples \cite{tuna2022exploiting} and quantified it by introducing \emph{perturbation} during inference time \cite{tuna2022exploiting, gal2016dropout, seebock2019exploiting, cremades2019reynolds}. However, there has been limited exploration of this uncertainty in the context of LLMs.


In this paper, we introduce a novel approach, Sampling with Perturbation for Uncertainty Quantification (SPUQ), as depicted in Fig.\ref{fig-flowchart}. SPUQ addresses both \emph{aleatoric} and \emph{epistemic} uncertainties in LLMs by amalgamating existing methods that assess uncertainty from disparate angles. Our method tackles \emph{epistemic} uncertainty through a perturbation module tailored for LLMs. This module, inspired by \cite{cremades2019reynolds, tuna2022exploiting}, gauges model sensitivity to input perturbations. Coupled with this is our approach to \emph{aleatoric} uncertainty, which adopts the principles of sampling methodologies \cite{wang2022self, si2022prompting}, enhanced by an aggregation module.

Within the perturbation module, we adjust temperature and/or prompts using a variety of techniques, such as paraphrasing, dummy tokens, and perturbed system messages. Our aggregation module not only generalizes the``exact match'' method used in existing sampling approaches \cite{wang2022self, si2022prompting} as a more general \emph{inter-sample} matching method but also incorporates \emph{intra-sample} metrics \cite{chen1998evaluation, lin2022teaching}. In essence, SPUQ distinguishes itself from traditional sampling-based UQ methods through its innovative perturbation and aggregation modules. The former ensures a comprehensive treatment of epistemic uncertainty, while the latter adapts the method to a wider array of text generation tasks.

We perform thorough experimental studies on different perturbation and aggregation techniques using multiple question-answering datasets for various LLMs, such as GPT and PaLM series \cite{chowdhery2022palm}. Through comparing our perturbation-based UQ method with existing baselines, we exhibit the efficiency of our approach in enhancing model uncertainty calibration, reducing Expected Calibration Error (ECE) by 50\% on average.

In summary, our key contributions in this work are threefold:
1) We introduce a novel perturbation sampling-based uncertainty quantification (SPUQ) framework tailored for LLMs. This framework effectively addresses both aleatoric and epistemic uncertainties, leading to improved model uncertainty calibration.
2) We unify and generalize existing methods, integrating them into our perturbation and aggregation modules, making them adaptable to a broad range of LLM tasks.
3) We demonstrate significant improvement in uncertainty calibration through comprehensive experimental studies on multiple datasets across different LLMs.

\section{SPUQ: Sampling with Perturbation for Uncertainty Quantification}

\begin{figure}[h]
\centering
\includegraphics[width=7cm]{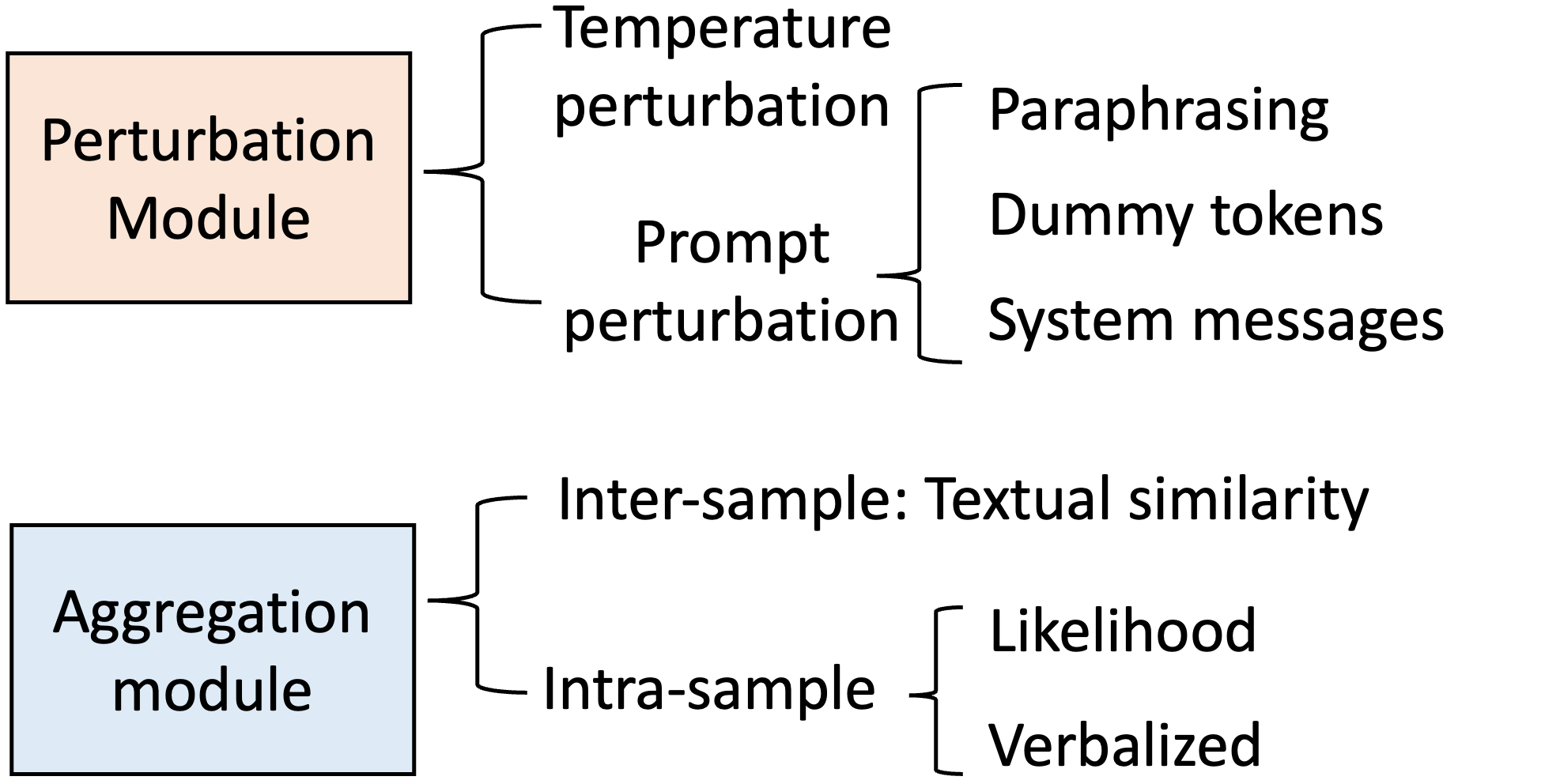}
\caption{\label{fig-hparam}
Options associated with the perturbation (Section~\ref{sec-perturb-module}) and aggregation modules (Section~\ref{sec-agg-module}) of the SPUQ method.
}
\end{figure}

In this section, we introduce the Sampling with Perturbation for Uncertainty Quantification (SPUQ) technique, as depicted in Fig.\ref{fig-flowchart} and elaborated in Algorithm \ref{algo-spuq}. SPUQ operates on an LLM's original input temperature $T_0$ and prompt $x_0$. It derives perturbed variants $(T_i, x_i)$ using the perturbation module and aggregates the outputs ${y_i}$ from both the original and perturbed LLM inputs to compute a confidence score, $c$. This score represents the quantified uncertainty, where, in our implementation, $c$ ranges from 0 to 1; a higher $c$ denotes reduced uncertainty. As previously highlighted, SPUQ addresses \emph{epistemic} uncertainties via its perturbation module and \emph{aleatoric} uncertainties through the sampling approach. We have proposed and evaluated diverse techniques for both the perturbation and aggregation modules, represented in Fig.\ref{fig-hparam}.

\begin{algorithm}
\caption{SPUQ}
\begin{algorithmic}[1]
\label{algo-spuq}
\REQUIRE 
Original temperature $T_0$, original prompt $x_0$, perturbation and aggregation module hyperparameters (Fig.~\ref{fig-hparam})
\ENSURE Confidence score $c$
\STATE Obtain $k$ perturbed inputs $(T_i, x_i)$ from the perturbation module, where $i = 1$ to $k$.
\FOR{$j = 0$ to $k$}
    \STATE Send $(T_j, x_j)$ to LLM being tested
    \STATE Obtain output $y_j$ from LLM
\ENDFOR
\STATE Send set $\{y_j\}$ to the aggregation module
\STATE Obtain the confidence score $c$
\RETURN $c$
\end{algorithmic}
\end{algorithm}

\subsection{Perturbation Module}
\label{sec-perturb-module}

\begin{table*}
\centering
\small
\begin{tabular}{p{0.6\linewidth}ll}
\toprule
\textbf{Input} $x_i$ & \textbf{Prediction} $y_i$ & \textbf{Likelihood} \\
\hline
\textbf{Original} \\
Will Jay-Z reach the age of 60 before Kendrick Lamar? & No \emph{(Incorrect)} & 92\% \\
\textbf{Perturbed prompt} \\
Is it likely that Jay-Z will turn 60 before Kendrick Lamar does? & Yes & 83\% \\
Will Jay-Z hit the age of 60 before Kendrick Lamar does? & No & 97\%\\
Before Kendrick Lamar, will Jay-Z reach the age of 60? & Yes & 94\% \\
Is it possible that Jay-Z will turn 60 before Kendrick Lamar? & No & 95\%\\
\bottomrule
\end{tabular}
\caption{\label{table-example}
A motivating example from the StrategyQA dataset. GPT-3 confidently predicts an incorrect answer with a high likelihood (92\%). In this case, uncertainty quantification relying solely on likelihood or re-sampling leads to overconfidence. However, input perturbation via paraphrasing reveals significant prediction instability, enabling the detection of epistemic uncertainties.
}
\end{table*}

To encapsulate epistemic uncertainties, our focus shifts towards understanding the susceptibility of model outputs to minor perturbations. Existing methods introduce perturbations using Monte Carlo Dropout \cite{seebock2019exploiting, tuna2022exploiting, gal2016dropout}. However, this approach is not feasible for closed LLM APIs. As an alternative, we introduce perturbations to the model inputs: the temperature and the prompt. 

For temperature perturbation, we either adopt a temperature deviating from \( T_0 \) consistently across all inputs, or we sample temperatures randomly\footnote{For GPT series, we sample temperature uniformly from 0.0 to 2.0, and for PaLM series, the range is restricted from 0.0 to 1.0 by the API.} to determine each \( y_i \).
When perturbing prompts, our objective remains consistent: introduce lexical variations without altering the core meaning. Our experiments encompass three strategies: 

\paragraph{Paraphrasing} We generate \( k \) paraphrased inputs, $\{x_i\} = \text{paraphraser}(x_0)$, for \( i = 1, \ldots, k \), by querying ChatGPT (\texttt{gpt-35-turbo-v0301}) using the following prompt:
\texttt{``
Suggest \{k\} ways to paraphrase the text in triple quotes above. If the original text is a question, ensure your suggestions retain a question. Provide your response in JSON format: \{"paraphrased": list of str\}
''}
It's noteworthy that this procedure involves only a single LLM call to obtain all \( k \) paraphrased prompts.

\paragraph{Dummy Tokens} We randomly select tokens, denoted by \( d \), that marginally influence the original meaning and prepend or append them to \( x_0 \). Such tokens could be newline characters, tab spaces, ellipses, or supplementary punctuation marks like extra question marks for queries. The altered prompts can be described as \( x_i=x_0 + d_i \) or \( x_i= d_i + x_0 \).

\paragraph{System Messages} For chat-mode LLM, such as ChatGPT, GPT-4, and PaLM2-Chat, perturbations can be introduced not just to the user prompt—which includes the actual query—but also to the system message. Given an original system message like \texttt{``You are a helpful assistant''}, we implement perturbations by replacing it with a randomly chosen message from a predefined set. Examples encompass phrases such as an empty system message, or semantically similar messages like \texttt{``You are a friendly assistant''}, \texttt{``You are a question-answering assistant''}, and \texttt{``You are a supportive question-answering assistant''}\footnote{Our current implementation and analysis may be somewhat constrained to these basic system messages. The ramifications of perturbations on more sophisticated and complex system messages warrant exploration in future works.}.

\subsection{Aggregation Module}
\label{sec-agg-module}

The vanilla sampling techniques \cite{si2022prompting, wang2022self} are proposed in scenarios where $\{y_i\}$ can be compared using the ``exact match'' criterion, which may not be suitable for a broader array of text generation tasks. To address this, we introduce an augmented method tailored for general language models, incorporating various aggregation methods (refer to Fig.~\ref{fig-hparam}) to derive the confidence score $c$ without necessitating an exact match.

\paragraph{Inter-sample} This approach revolves around the textual similarity among sample outputs \( \{y_i\} \).

\begin{equation}
c_\text{inter} = \frac{ \sum^k_{i=0, i\neq j} s(y_j, y_i) w_i } { \sum^k_{i=0, i\neq j} w_i }
\label{eq:weighted}
\end{equation}

Here, \( j \) signifies the index of the output being assessed for accuracy. The function \( s(y_j, y_i) \) measures the textual similarity between outputs \( y_i \) and \( y_j \), and \( w_i \) designates the weight allocated to the output variant \( y_i \). 
Vanilla sampling's majority-voting approach \cite{si2022prompting, wang2022self} becomes a specific instance of this formula if \( j \) is set to the most frequent answer, \( s(y_j, y_i) \) is configured as the exact match function (yielding a value of 1 exclusively when \( y_i \) precisely aligns with \( y_j \)), and \( w_i=1 \) uniformly for all \( i \). 

To calibrate uncertainty relative to the accuracy of original prompts, we assign \( j=0 \). For perturbed prompts, we determine the weight \( w_i \) as \( s(x_0, x_i) \), signifying the similarity between the perturbed prompt \( x_i \) and the original \( x_0 \). This configuration prioritizes milder perturbations over extreme ones, mitigating potential repercussions from severe perturbations, such as unsatisfactory paraphrasing.

Regarding the similarity function \( s(\cdot,\cdot) \), our experiments span three metrics: \texttt{BERTScore}\footnote{\url{https://huggingface.co/spaces/evaluate-metric/bertscore}} \cite{zhang2019bertscore}, cosine similarity derived from \texttt{SentenceBERT} embeddings\footnote{\url{https://www.sbert.net}} \cite{reimers2019sentence}, and the \texttt{RougeL} Score\footnote{\url{https://github.com/google-research/google-research/tree/master/rouge}} \cite{lin2004rouge}.

\paragraph{Intra-sample} The second category computes the average of the uncertainties discerned individually for each sample output, $c(x_i, y_i)$. We employ two strategies to obtain $c(x_i, y_i)$. First, we utilize likelihood (or the reciprocal of perplexity \cite{chen1998evaluation}). This approach, however, is confined to LLM APIs that grant access to the predicted token distribution. Subsequently, we adopt the verbalized uncertainty strategy posited by \citet{lin2022teaching}. Post-generation of the output \( y_i \), we prompt the LLM to articulate its uncertainty, either as words or as numbers, with the prompts listed in Table~\ref{table-verbalized} in the \emph{Appendix}.

\begin{equation}
c_\text{intra} = \frac{ \sum^k_{i=0} c(x_i, y_i)} { k + 1}
\label{eq:weighted}
\end{equation}

\subsection{Selecting Hyperparamters}
\label{sec-tune}

\begin{figure*}[ht]
\centering
\includegraphics[width=16cm]{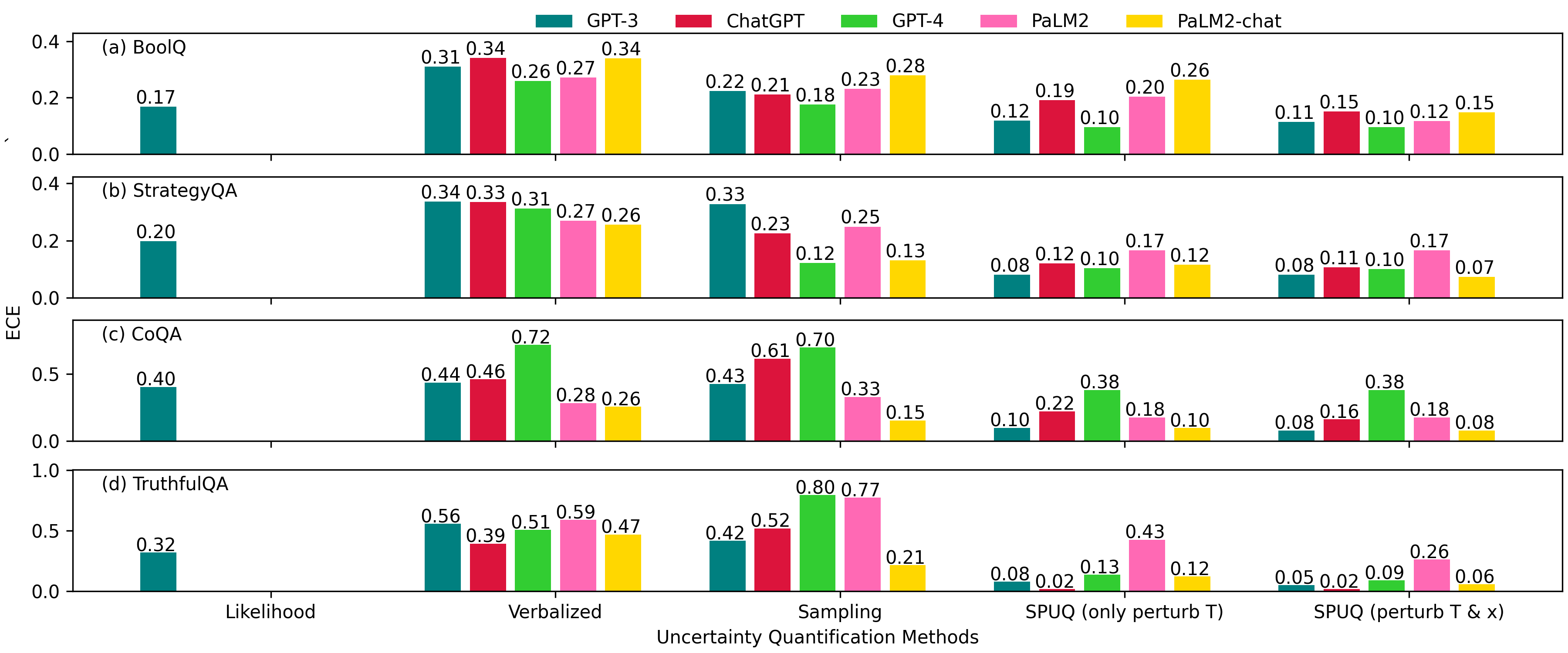}
\caption{\label{fig-eval}
An overview of the uncertainty calibration performance, measured by the Expected Calibration Error (ECE), for various uncertainty calibration methods across five LLMs over four question-answering datasets. A lower ECE indicates better uncertainty calibration. 
}
\end{figure*}

\begin{table*}
\centering
\small
\begin{tabular}{c|ccc|ccc}
\toprule 
 \multirow{2}{*}{Dev set} & \multicolumn{3}{c|}{Tuned SPUQ hyperparameters} & \multicolumn{2}{c}{Test performance} \\
 & $T$ perturbation & $x$ perturbation & Aggregation & ECE $\downarrow$ & Correlation $\uparrow$ \\
 \hline
I & +0.3 & Paraphrasing & inter-sample, RougeL & 0.082 & 0.690 \\
II & +1.3 & Paraphrasing & inter-sample, RougeL & 0.102 & 0.670 \\
III & +0.3 & Paraphrasing & inter-sample, RougeL & 0.082 & 0.690 \\
V & +1.3 & Dummy tokens & inter-sample, RougeL & 0.103 & 0.672 \\
IV & +0.3 & Paraphrasing & inter-sample, RougeL & 0.082 & 0.690 \\
\bottomrule
\end{tabular}
\caption{\label{table-tune}
Hyperparameters tuned on five separate development sets based on best ECE for GPT-4 on the TruthfulQA dataset and base value $T_0=0.7$. 
Test performance is reasonably robust to the choice of development set. 
}
\end{table*}

The optimal selection of specific perturbation and aggregation methods may vary depending on the dataset and LLM in use. We employ a development set to fine-tune these hyperparameters based on ECE and find that performance remains reasonably robust regardless of the development set chosen (refer to Table~\ref{table-tune} for an example). Furthermore, through empirical investigations spanning five LLMs and four question-answering datasets, we note that certain perturbation and aggregation methods consistently yield better calibration results than others (refer to Section~\ref{sec-hparam}). This offers guidance on the recommended choices for these hyperparameters.

\section{Experimental Setup}

\paragraph{Large Language Models (LLMs)} We conduct experiments using five LLMs\footnote{GPT series are accessed via Azure OpenAI API. PaLM series are accessed via Google Cloud Platform API.}: GPT-3 \cite{brown2020gpt3} (\texttt{text-davinci-003}), ChatGPT \cite{chatgpt} (\texttt{gpt-35-turbo-v0301}), GPT-4 \cite{openai2023gpt4} (\texttt{gpt-4}), PaLM2 \cite{chowdhery2022palm} (\texttt{text-bison}), and PaLM2-Chat \cite{chowdhery2022palm} (\texttt{chat-bison}).

\paragraph{Datasets} We evaluate our methods on a summarization task dataset, XSUM \cite{narayan2018xsum}, and four question-answering datasets: 
\begin{enumerate}
    \item \emph{Classification-type} datasets, which include two binary (yes/no) sets: {StrategyQA}~\cite{geva2021strategyqa} and {BoolQ}~\cite{clark2019boolq}.
    \item \emph{Generation-type} datasets, featuring the {CoQA}~\cite{reddy2019coqa} and {TruthfulQA}~\cite{lin2021truthfulqa} collections.
\end{enumerate}

\paragraph{Baselines} Our SPUQ method is benchmarked against several established baselines: 
\begin{itemize}
    \item Likelihood \cite{chen1998evaluation}, where the confidence score is defined as the length-normalized LM likelihood. Only applicable to GPT-3 due to API constraints.
    \item Verbalized \cite{lin2022teaching} method (refer to Table~\ref{table-verbalized} for the prompts we employed).
    \item Sampling without perturbation \cite{si2022prompting}. To adapt this method to generation-type datasets, we substitute the exact match criterion with textual similarity\footnote{For fair comparison, we tune on a development set to find the best text similarity metric for this baseline.}.
\end{itemize}

\paragraph{Evaluation} The calibration quality of the uncertainty is assessed using a set of metrics: Expected Calibration Error (ECE)\footnote{ECE calculated as the mean of the absolute difference between each confidence bucket's accuracy and average confidence score $c$} following \cite{si2022prompting, openai2023gpt4} and the Pearson's correlation ($\rho$) between confidence score $c$ and accuracy.
LLM accuracy is assessed via the ``exact match'' criterion for classification-type datasets, and F1 criterion for generation-type datasets following \cite{reddy2019coqa}. 
As mentioned in Section~\ref{sec-tune}, hyperparameters for perturbation and aggregation may be tuned on a development set. To examine its sensitivity to development sets, for SPUQ, we report the average evaluation results of five tuning runs. For each run, hyperparameters are selected based on ECE on a development set of 30 randomly selected samples.

\section{Results and Discussion}

\begin{figure}[h]
\centering
\includegraphics[width=8cm]{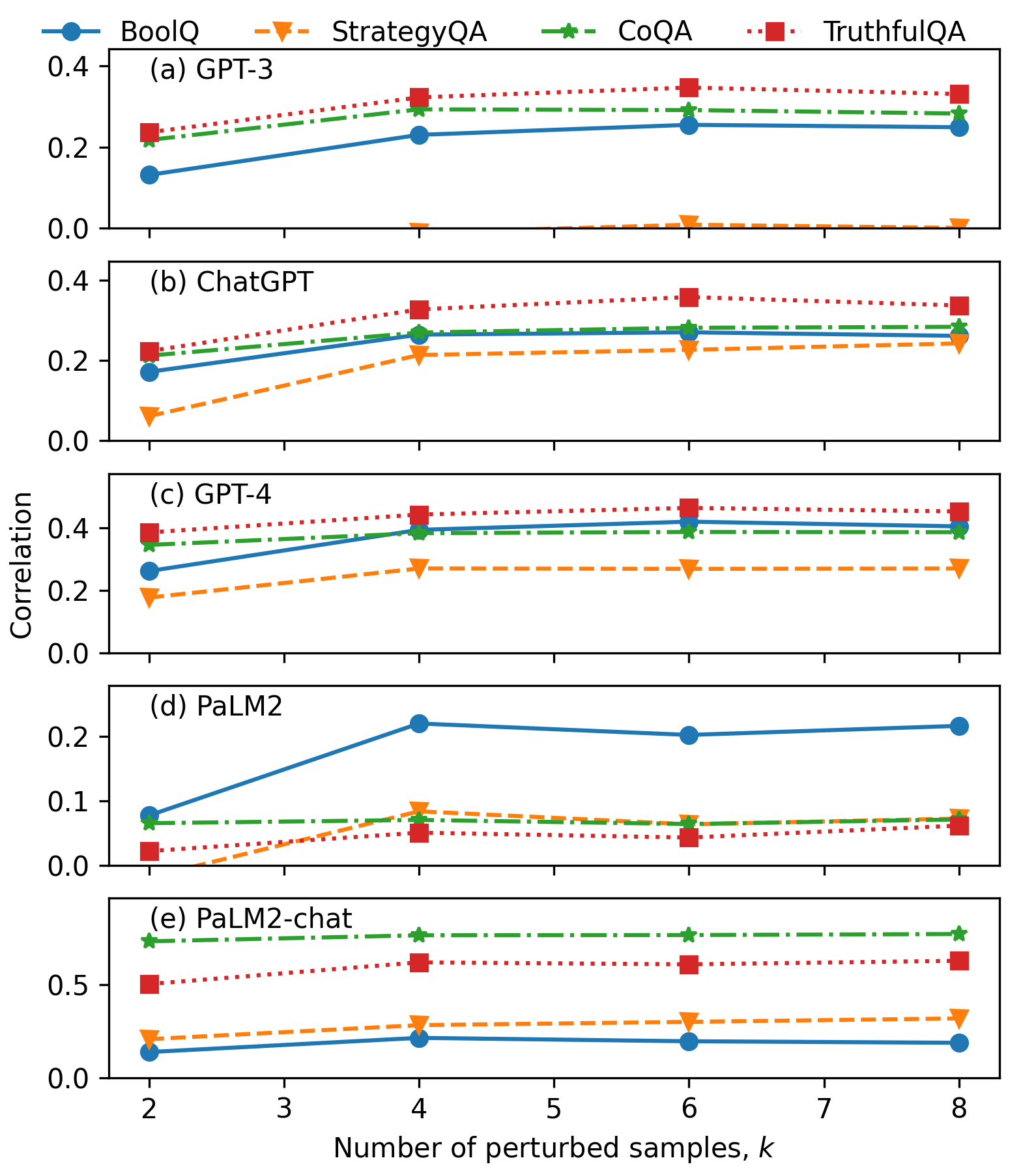}
\caption{\label{fig-n-ens}
The dependence of the uncertainty calibration, measured by the average confidence-accuracy Pearson correlation, on the number of perturbed samples, $k$.
The general trend indicates that calibration improves as $k$ increases, but it plateaus approximately at $k$=5.
}
\end{figure}

\begin{figure}[h]
\centering
\includegraphics[width=6cm]{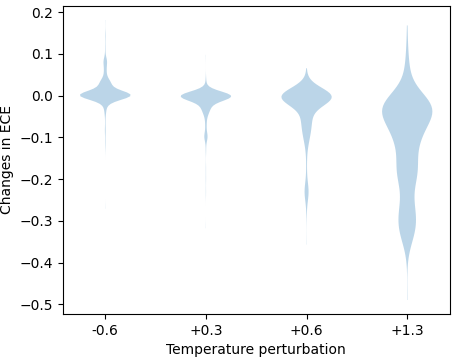}
\caption{\label{fig-T-voilin}
The distribution of ECE changes for specific temperature perturbations, taking into account variations in other hyperparameters. An increase in temperature (base value is $T_0=0.7$) during the sampling process tends to enhance calibration (decreased ECE).
}
\end{figure}

\subsection{Enhanced Uncertainty Calibration}

\paragraph{Overview}
Our results, depicted in Fig.\ref{fig-eval} and Table~\ref{table-xsum}, highlight SPUQ's efficacy: it consistently posts the lowest ECE across most tested language models and datasets, outperforming baselines with a 50\% reduction on average (30\% to 70\% depending on the dataset and LLM). This superior calibration also aligns with the observed correlation between confidence and accuracy, detailed in Fig.~\ref{fig-eval-corr} in the \emph{Appendix}.
The enhancement achieved by SPUQ, when compared to the sampling approach without perturbation, suggests that the improvement is primarily due to the perturbation module, specifically designed to address epistemic uncertainty. Notably, both temperature and prompt perturbations play significant roles in this enhancement.

\begin{table}[ht]
\centering
\small
\begin{tabular}{cccc p{4cm}}
\toprule
LLM & Likelihood & Sampling & SPUQ \\
\hline
GPT-3 & 0.386 & 0.393 & 0.214 \\
ChatGPT & - & 0.406 & 0.209 \\
\bottomrule
\end{tabular}
\caption{\label{table-xsum}
ECE on the XSUM summarization task.
}
\end{table}

\paragraph{A Case Study}
Table~\ref{table-example} elucidates the impact of perturbation. Occasionally, LLMs may produce confidently erroneous predictions. In the given example, there's a striking 92\% likelihood of generating the incorrect response "No". Utilizing conventional sampling with unaltered input parameters \cite{si2022prompting}, the model typically mirrors the initial output, yielding an overconfident score close to 1, which results in poor uncertainty calibration. In contrast, paraphrasing the prompt brings forth epistemic uncertainty. Even minor changes can lead to starkly different outputs. For instance, in our example, the probabilities associated with "Yes" and "No" fluctuate markedly across the five $y_i$. Thus, SPUQ proves more discerning, delivering a confidence score of 0.50 since only half of the sampled $y_i$ matches $y_0$

\begin{figure*}[h]
\centering
\includegraphics[width=13.55cm]{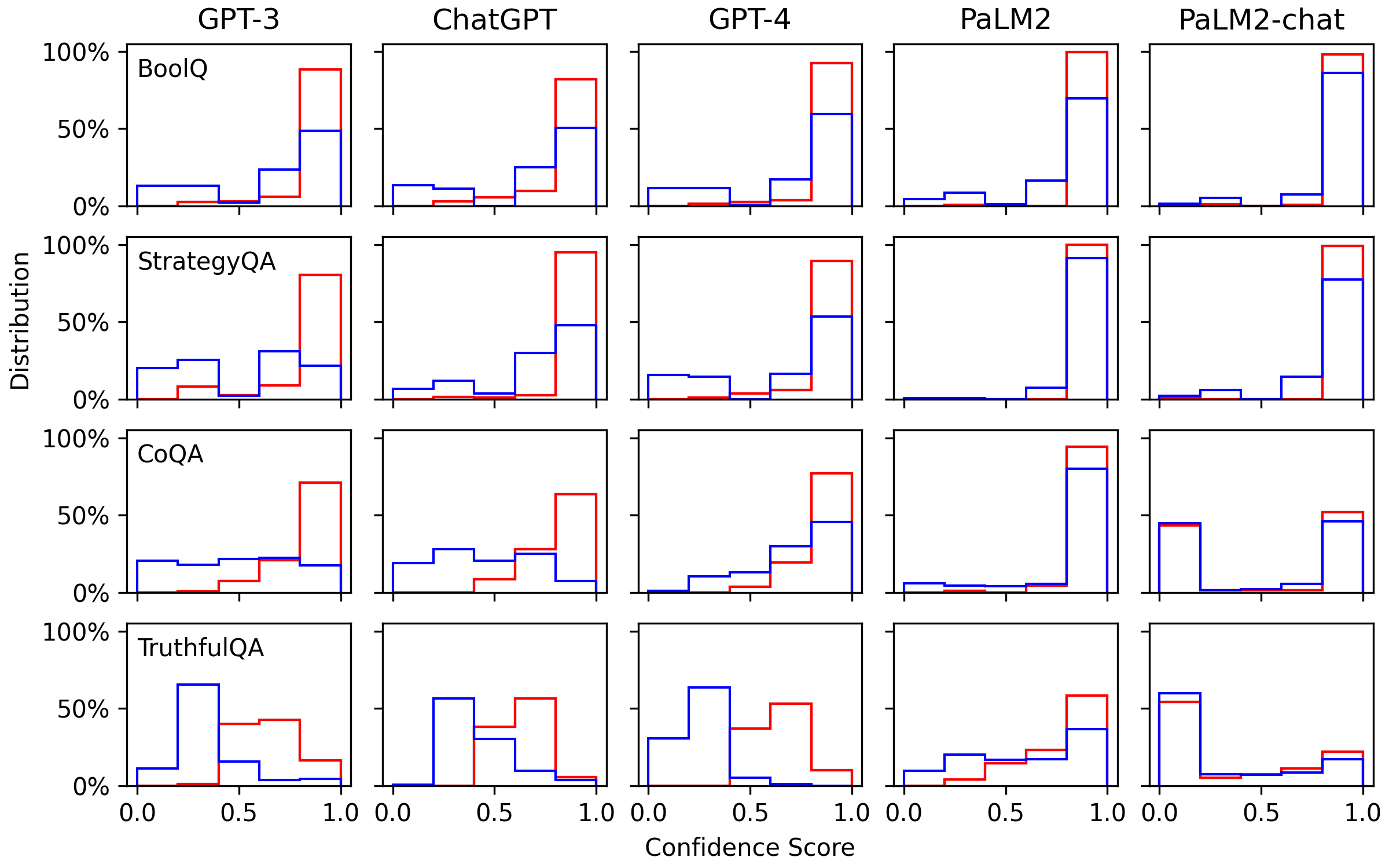}
\caption{\label{fig-conf-hist}
The empirical distribution of the confidence scores obtained using the conventional sampling (red) and our SPUQ (blue) approach. SPUQ displays a flatter distribution and less frequent over-confidence. 
}
\end{figure*}


\paragraph{Mitigating Overconfidence}
Our case study highlights SPUQ's capacity to measure epistemic uncertainties by exploiting prediction instability through input perturbations. This is expected to temper the overconfidence frequently displayed by LLMs. Notably, confidence score distributions from SPUQ demonstrate a more even spread than those obtained via unperturbed sampling, as showcased in Figure~\ref{fig-conf-hist}.

\subsection{Dependence on Hyperparameters}
\label{sec-hparam}

In this subsection, we delve into how calibration is influenced by various hyperparameters.

\paragraph{Number of Perturbed Samples} Intuitively, increasing the number of samples should render the uncertainty quantification more consistent and trustworthy, eventually converging to a specific value. Our empirical findings affirm this hypothesis. As delineated in Fig.~\ref{fig-n-ens}, the overarching trend signifies that calibration progressively refines with an upsurge in the number of perturbed samples \( k \). However, the improvement begins to level off, approximately at \( k = 5 \).

\paragraph{Temperature Perturbation} 
By sampling at a heightened temperature, the predicted distribution leans towards uniformity, making the generated outcomes more disparate from the original output. Consequently, the sampling method becomes less skewed by overconfidence, enhancing uncertainty calibration. This dynamic is visually corroborated in Fig.~\ref{fig-T-voilin}. As the temperature during the sampling procedure escalates, ECE diminishes, signaling superior calibration.  \footnote{Our observations did not show improvements in calibration with random temperature perturbation, suggesting that directional adjustments (specifically increasing the temperature) are more effective.}

\begin{figure}[hb]
\centering
\includegraphics[width=7.5cm]{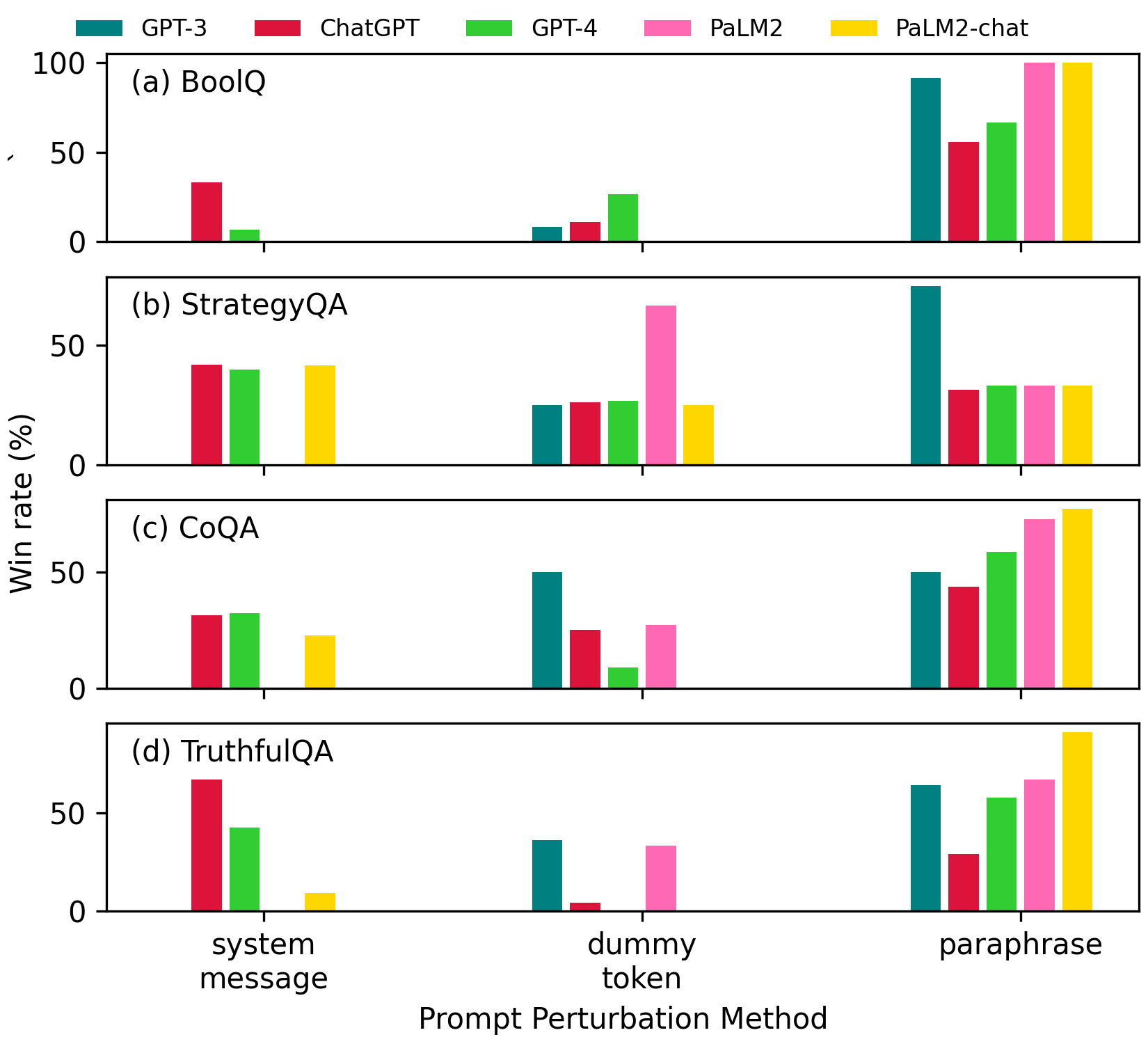}
\caption{\label{fig-para}
Dependence of uncertainty calibration on the prompt perturbation method. The "win rate" indicates the percentage to achieve the lowest ECE against others. 
}
\end{figure}

\paragraph{Prompt Perturbation} While each of the three prompt perturbation methods we experimented with is designed to retain the original meaning, they manifest distinct perturbation characteristics. Paraphrasing can modify a significant portion of the input tokens. In contrast, dummy tokens and system messages do not make any changes to the question being asked.
Fig.~\ref{fig-para} shows that the paraphrasing method demonstrates superior calibration in more than half of the test cases, compared to the other prompt perturbation methods.

\begin{figure}[h!]
\centering
\includegraphics[width=6.5cm]{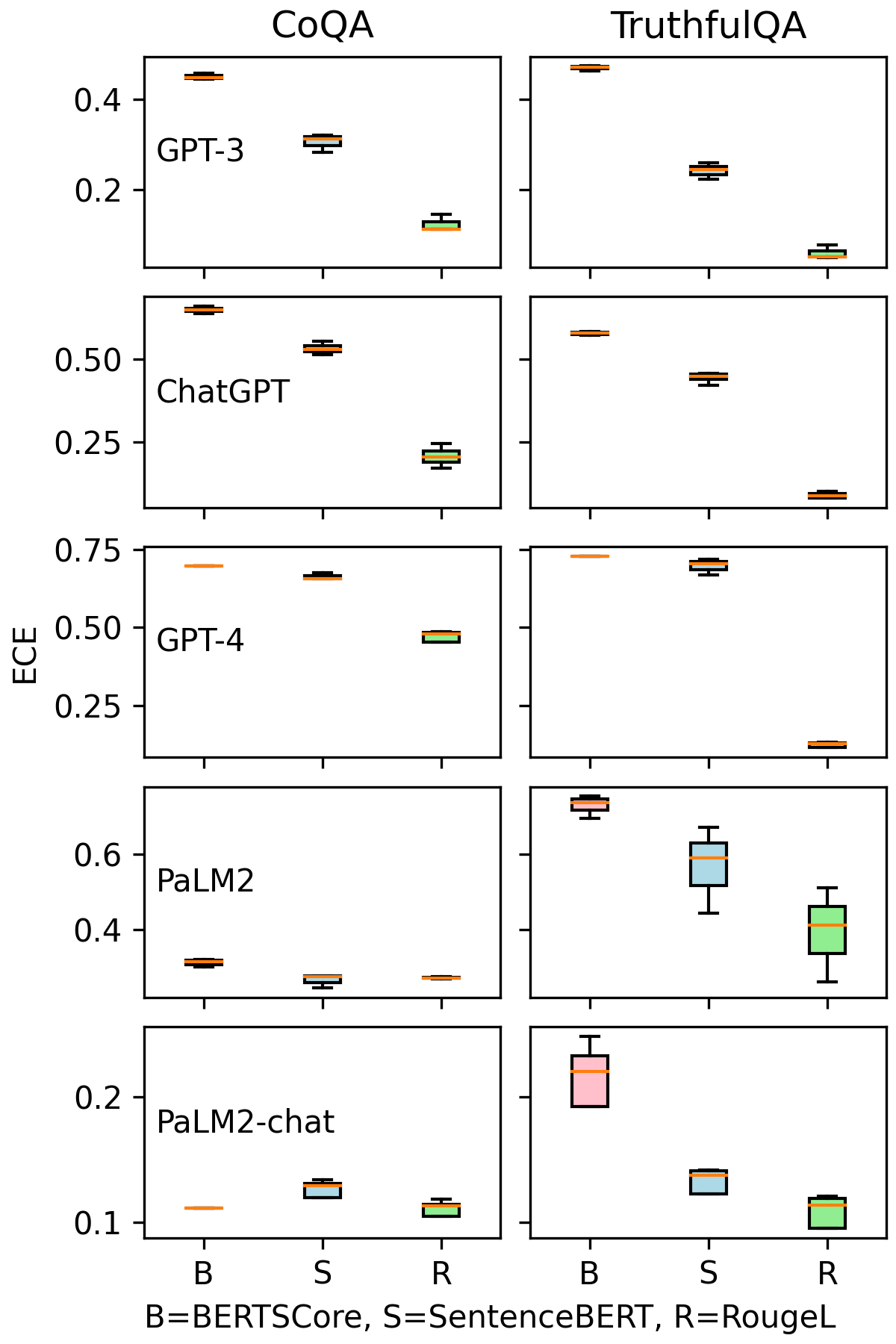}
\caption{\label{fig-sim}
The distribution of ECE on various text similarity metrics employed by SPUQ. 
}
\end{figure}

\paragraph{Aggregation Module} While the inter-sample aggregation method has been employed in prior work \cite{si2022prompting}, we have further generalized it by incorporating textual similarity. Among the three similarity metrics assessed\footnote{BoolQ and StrategyQA are only assessed by ``exact match'' as they are binary (Yes/No) question answering}, the RougeL score consistently outperforms the other two, namely BERTScore and SentenceBERT, as shown in Fig.~\ref{fig-sim}. Interestingly, our findings also indicate that the inter-sample aggregation isn't the sole viable approach; the intra-sample method also emerges as a compelling choice. No single aggregation method distinctly surpasses the others in all test scenarios (Refer to Fig.~\ref{fig-agg} in the \emph{Appendix} for details). Consequently, we advocate for experimentation with both inter and intra-sample aggregation methods, rather than exclusively adhering to the inter-sample approach prevalent in existing literature.

\paragraph{Robustness to the Development Set} As previously mentioned, we repeat the hyperparameter tuning process five times to assess its robustness to the choice of the development set. Notably, test outcomes remain robust across different development sets, with ECE standard deviation being roughly 10\% of the mean—minor when contrasted with the 30\% to 70\% improvement. Table~\ref{table-tune} offers a sample of these tuning results.

\section{Related works}


\paragraph{Hallucination} Hallucination in LLMs is a significant challenge, as it can be induced by data, training, and inference processes \cite{ji2023hallucination}. Detecting hallucination on-the-fly remains a daunting task. Language models tend to over-commit to early mistakes, leading to more errors and contributing to hallucination snowballing \cite{zhang2023snowball}. 

\paragraph{Improving Reliability} Techniques to improve LLMs' reliability and reduce uncertainty have been proposed in the literature \cite{zhou2023navigating}. Wang et al. \cite{wang2022self} found that sampling and aggregating multiple chain-of-thought reasoning paths can enhance LLMs' performance and reliability. Good in-context prompting strategies, such as few-shot prompting, can improve GPT-3's reliability\cite{si2022prompting}.

\paragraph{Uncertainty Quantification} UQ in deep learning models has been explored using various techniques, such as Bayesian approximation and ensemble learning \cite{abdar2021review, malinin2020uncertainty}. 
LLMs are prompted to self-evaluate their previous predictions \cite{kadavath2022language} or to express their uncertainty in natural language \cite{lin2022teaching}. On the other hand, sampling-based methods \cite{si2022prompting} like Semantic Uncertainty \cite{kuhn2023semantic} consider linguistic invariances when quantifying uncertainty. 

\section{Conclusion}

We introduced the SPUQ method to enhance uncertainty calibration in LLMs, achieving a notable reduction in Expected Calibration Error by 30\% to 70\%. Our ablation study linked this improvement largely to our perturbation mechanism, underscoring its role in addressing epistemic uncertainty. The application of SPUQ offers a path to more reliable LLM outputs. Future work should expand SPUQ's applicability on beyond present datasets with simple prompts, exploring its effectiveness across diverse tasks and complex prompt structures.

\section*{Limitations}

We experimented with datasets where accuracy can be assessed relatively easily with the reference answer. Future works are encouraged on tasks where accuracy is less well defined, such as in conversation and content generation.
Our approach introduces additional computational costs due to multiple generations and/or paraphrasing, which may increase the latency of the output. The steps, however, can be parallelized to ensure $\mathcal{O}(1)$ complexity.

\section*{Ethics Statement}

This work quantifies and reduces uncertainty in LLM outputs. It may help to reduce the generation and use of mistaken or misleading content from LLMs, to encourage a safer use of these models.

\clearpage

\bibliography{custom}
\bibliographystyle{acl_natbib}

\clearpage

\appendix

\section{Appendix}
\label{sec:appendix}

\begin{figure*}[b]
\centering
\includegraphics[width=16cm]{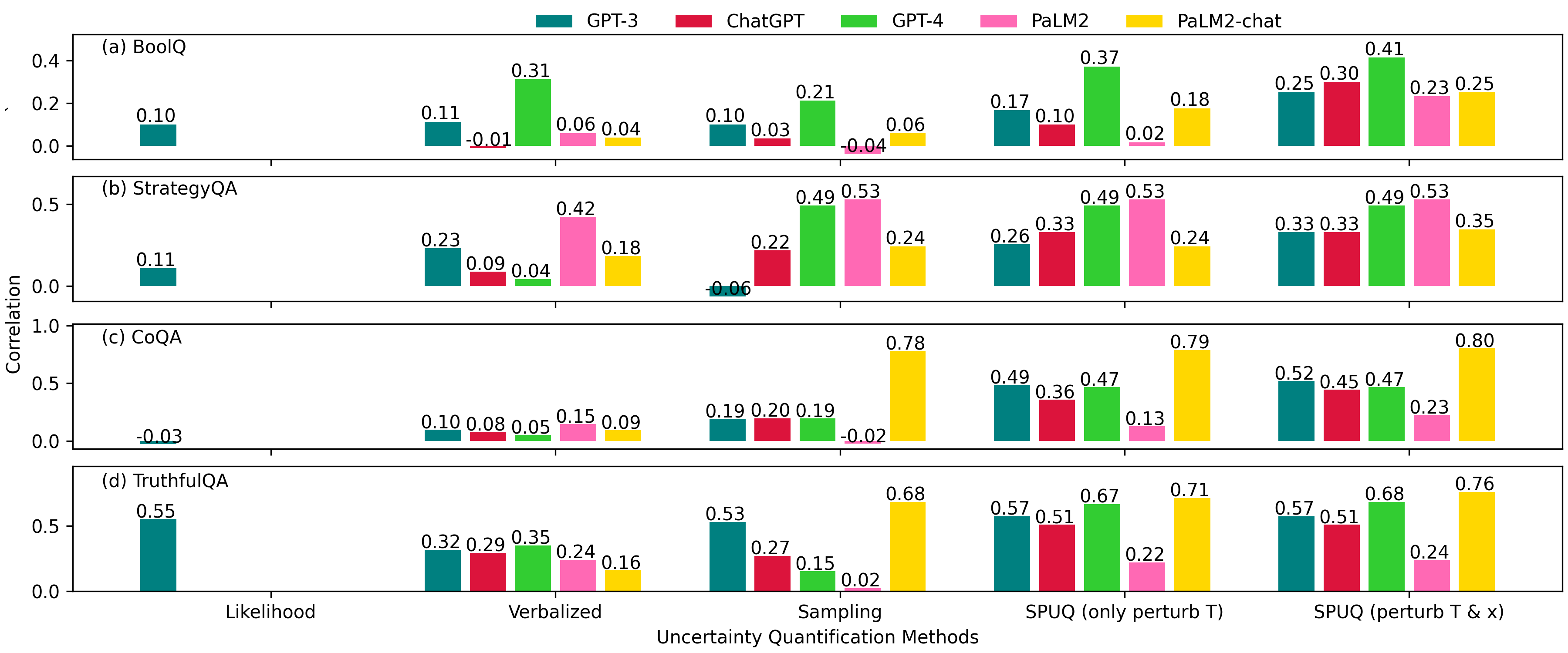}
\caption{\label{fig-eval-corr}
An overview of the uncertainty calibration performance, measured by, $\rho$, the Pearson's correlation between confidence score $c$ and accuracy, for various uncertainty calibration methods across five LLMs over four question-answering datasets. A higher $\rho$ indicates better uncertainty calibration. Our method, sampling with perturbation, exhibits the highest $\rho$ for a given LLM in most cases.
}
\end{figure*}

\begin{figure}[ht]
\centering
\includegraphics[width=8cm]{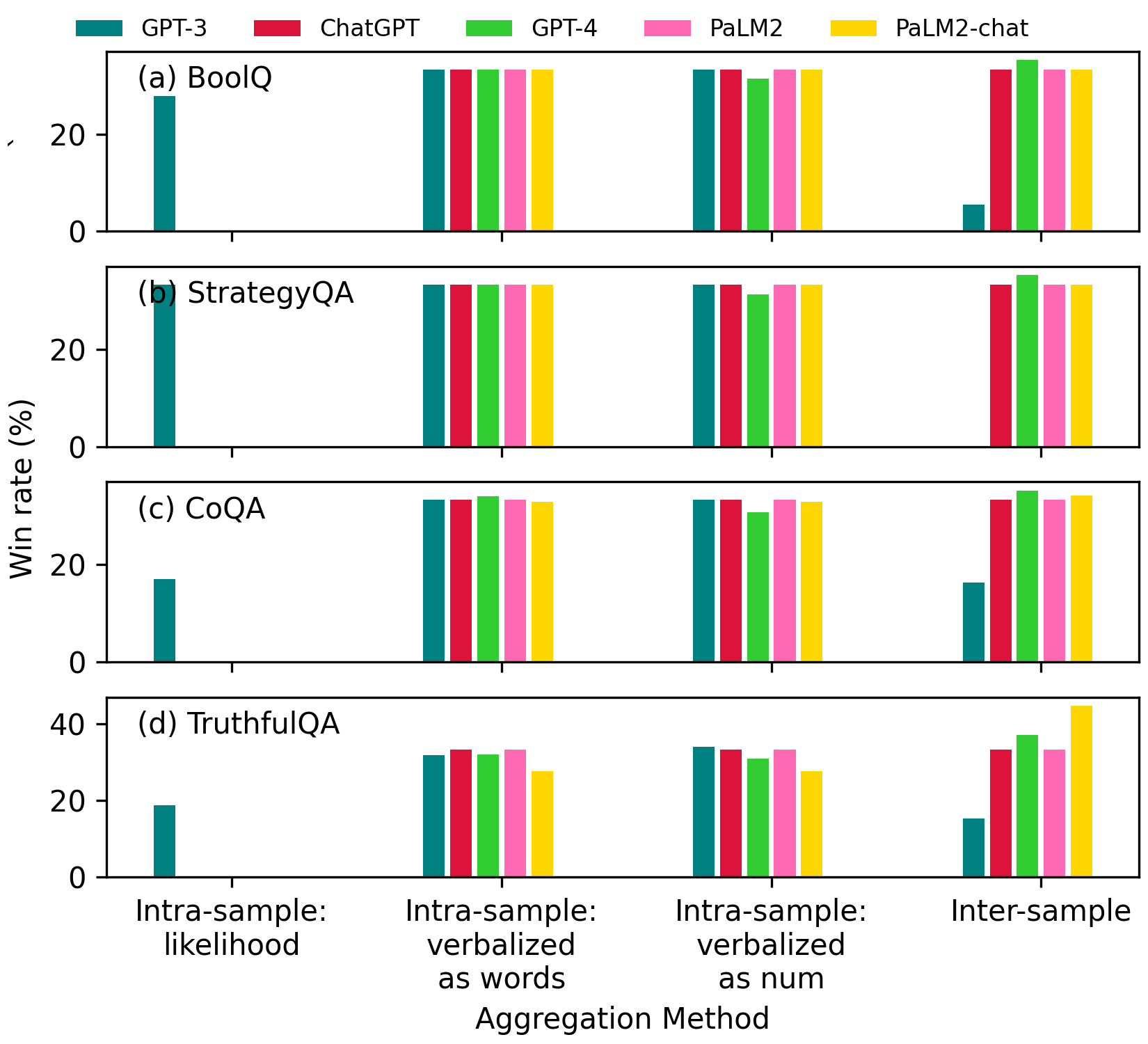}
\caption{\label{fig-agg}
Dependence of uncertainty calibration on the aggregation method. The "win rate" indicates the percentage of instances a method achieves the lowest ECE against others. 
}
\end{figure}

In the appendix, we include the results for the aggregation method (Fig.~\ref{fig-agg}), the overall evaluation using accuracy-confidence correlation (Fig.~\ref{fig-eval-corr}), and the prompts we used to obtain verbalized uncertainty (Table~\ref{table-verbalized}).

\begin{table}[ht]
\centering
\small
\begin{tabular}{llp{4cm}}
\toprule
Verbalized & LLM & Prompt \\
\hline
\multirow{2}{*}{Words} & chat & Your confidence is? (low, medium, high) \\
 & text & Confidence (low, medium, high): \\
\hline
\multirow{2}{*}{Numbers} & chat & Your confidence is? (a score between 0.0 to 1.0) \\
 & text & Confidence (a score between 0.0 to 1.0): \\
\bottomrule
\end{tabular}
\caption{\label{table-verbalized}
Prompts employed to derive verbalized confidence \citet{lin2022teaching}, used for aggregating intra-sample uncertainty. For uncertainty verbalized as words, we set $c=0.25$ for ``low'', 0.5 for ``medium'', 0.75 for ``high''
}

\end{table}


\end{document}